\documentclass[review]{elsarticle}

\usepackage{lineno,hyperref}
\modulolinenumbers[5]

\usepackage{amsmath}
\usepackage{amssymb}
\usepackage{amsthm}
\usepackage{color}
\usepackage{subfig}
\usepackage{algorithm}
\usepackage{algorithmic}

\newcommand{\be}{\begin{equation}}
\newcommand{\ee}{\end{equation}}

\DeclareMathOperator*{\argmin}{\mathrm{argmin}}

\usepackage[normalem]{ulem}
\usepackage{soul}
\newcounter{ToDo}
\newcounter{guocomm}
\newcounter{Note}
\definecolor{blue-violet}{rgb}{0.54, 0.17, 0.89}
\definecolor{mygreen}{rgb}{0.0, 0.5, 0.0}
\definecolor{awesome}{rgb}{1.0, 0.13, 0.32}
\definecolor{yellow}{rgb}{0.8, 0.8, 0.0}
\definecolor{bostonuniversityred}{rgb}{0.8, 0.0, 0.0}

\journal{Neurocomputing}

\begin{document}

\begin{frontmatter}

\title{Collaborative Low-Rank Subspace Clustering}

\author[csu]{Stephen Tierney\corref{mycorrespondingauthor}}
\cortext[mycorrespondingauthor]{Corresponding author}
\ead{stierney@csu.edu.au}

\author[wsu]{Yi Guo}
\ead{y.guo@westernsydney.edu.au}

\author[usyd]{Junbin Gao}
\ead{junbin.gao@sydney.edu.au}

\address[csu]{School of Computing and Mathematics, Charles Sturt University, Bathurst, NSW 2795, Australia}
\address[usyd]{Discipline of Business Analytics, The University of Sydney Business School,\\ The University of Sydney, NSW 2006, Australia}
\address[wsu]{Centre for Research in Mathematics, School of Computing, Engineering and Mathematics, Western Sydney University, Parramatta, NSW 2150, Australia}

\begin{abstract}
In this paper we present Collaborative Low-Rank Subspace Clustering. Given multiple observations of a phenomenon we learn a unified representation matrix. This unified matrix incorporates the features from all the observations, thus increasing the discriminative power compared with learning the representation matrix on each observation separately. Experimental evaluation shows that our method outperforms subspace clustering on separate observations and the state of the art collaborative learning algorithm.
\end{abstract}

\end{frontmatter}

\section{Introduction}

A fundamental step in machine learning and data analysis in general is to reduce the dimension of the original data to facilitate further computation by removing irrelevant information and to reduce effects such as the curse of dimensionality \cite{bellman1957dynamic, friedman2001elements}. Dimensionality reduction is most commonly performed by Principal Components Analysis (PCA) \cite{jolliffe2002principal, RamsaySilverman2005, jacques2014functional}, which takes a collection of data points from their original high dimensional space and fits them to a single lower dimensional subspace. In reality however, large datasets are composed of a union or ensemble of multiple subspaces. Furthermore the subspace labels are unknown ahead of time we are presented with the two-step problem of subspace identification and fitting. The former step is known as subspace clustering.

Dimensionality reduction remains a requirement for machine learning despite increases in computational power and memory since the resolution of capture devices in all fields continues to increase. In particular dimensionality reduction is important in many Computer Vision applications due to the intrinsic high dimensional nature of of the data, since images can contain between 10 to 100 million pixels. There are many circumstances in Computer Vision under which the union of subspaces model holds and subspace clustering still remains as state of the art. The two prominent examples are identifying individual rigidly moving objects in video \cite{tomasi1992shape, costeira1998multibody, kanatani2002motion, jacquet2013articulated} and identifying face images of a subject under varying illumination \cite{basri2003lambertian, georghiades2001few}. In both these applications the subjects lie nicely on very low dimensional subspaces i.e. 4 and 9 respectively \cite{elhamifar2012sparse}. However they still present a challenge as the data can contain significant noise and corruptions. Subspace clustering has found many other computer vision applications such as image compression \cite{hong2006multiscale}, image classification \cite{zhang2013learning, DBLP:conf/dicta/BullG12}, feature extraction \cite{liu2012fixed, liu2011latent}, image segmentation \cite{yang2008unsupervised, cheng2011multi} and hand-written character clustering \cite{hastie1998metrics, elhamifar2012sparse}.

Subspace Clustering has been used outside of computer vision with success. For example it has found use in biological applications. DNA microarrays are a powerful tool in exploring gene expression and genotyping. In general the process requires counting the expression levels of genes. This process can be performed  by subspace clustering  \cite{parsons2004subspace, wallace2015application, wang2004fast, tchagang2014subspace, parvaresh2008recovering}. At a more human scale subspace clustering has been used to automatically segment human activities from motion capture data, with the assumption that each motion class forms a subspace \cite{zhu2014complex, yang2008distributed, li2015temporal}.

\section{Preliminaries}

Subspace clustering is the unsupervised classification of data points, in a set, into their respective subspace. We assume that there are $p$ subspaces $S_1, S_2,\dots, S_p$ with respective dimensions $\{d_i\}^p_{i=1}$. Both the number of subspaces $p$ and the dimension of each subspace $d_i$ are unknown. The point set $\mathcal X \subset \mathbb R^D$ of cardinality $N$ can be partitioned as $\mathcal X = \mathcal X_1 \cup \mathcal X_2 \cup \dots \cup \mathcal X_p$ where each $\mathcal X_i$ are the data points from subspace $i$.

It is common to assume that each of the data points are generated using the following additive model
\begin{align}
\mathbf {x = a + n}
\label{intro:data_model}
\end{align}
where $\mathbf x$ is the observed data point, $\mathbf a$ is the true data point lying perfectly on a subspace and $n$ is an independent and random noise term. It is rarely the case that $\mathbf X$ is noise or corruption free. The data is often subject to noise or corruption either at the time of capture (e.g. a digital imaging device) or during transmission (e.g. wireless communication). Two common noise assumptions for the noise type are Gaussian or Sparse noise. It is difficult to isolate the original data $\mathbf a$ from the noise $\mathbf a$ (unless the underlying structure is exactly known \cite{candes2011robust}).

Moving from an abstract set notation we can express the problem in more concrete terms by matrix notation. The observed data matrix $N$ observed column-wise samples $\mathbf X = [\mathbf x_1, \mathbf x_2, \dots, \mathbf x_N ]$  $\in \mathbb{R}^{D \times N}$. The objective of subspace clustering is to learn the corresponding subspace label $\mathbf l = [ l_1, l_2, \dots, l_N ] \in \mathbb{N}^{N}$ for each data point where $l_i\in \{1,\ldots,k\}$. 

The dominant subspace clustering algorithms fall under the class of Spectral Clustering and they are our primary concern in this paper. Spectral Clustering Algorithms are defined by their used of spectral clustering to segment the data using an affinity matrix, which encodes the similarity between the data points. Spectral clustering is a popular approach to general clustering problems since it is relatively simple, computationally inexpensive, is straight forward to implement and can out perform more traditional clustering techniques \cite{von2007tutorial}. They have also, in recent years, dominated the field of subspace clustering. This is because they do not increase in complexity with the number or dimension of subspaces, they are often more robust to noise or outliers and finally they provide a simple to understand work pipeline that is easily adapted and modified.

In general, spectral subspace clustering algorithms construct an affinity matrix $\mathbf W \in \mathbb R^{N \times N}$, where entry $W_{ij}$ measures the similarity between points $i$ and $j$. The hope is that $W_{ij} \gg 0$ when points $i$ and $j$ are in the same subspace and $W_{ij} = 0$ when they are not. Spectral clustering then constructs the  segmentation of the data is obtained by applying K-means to the eigenvectors of a Laplacian matrix $\mathbf L \in \mathbb R^{N \times N}$. In practice, Normalised Cuts (N-Cut) \cite{shi2000normalized}, is most often used to obtain final segmentation. N-Cut essentially follows the procedure outlined above. N-Cut has been shown to be robust in subspace segmentation tasks and is considered state of the art. In cases where N-Cut is too slow one can use approximate techniques such as the Nystr\"{o}m method \cite{fowlkes2004spectral}.

The differences in Spectral Clustering algorithms lies in the construction of $\mathbf W$. One cannot directly use the distance between data points since they may be close in the ambient space but lie along different subspaces as may occur at the intersection of subspaces. Conversely data points may be distant but still in the same subspace. 

Sparse Subspace Clustering (SSC) \cite{elhamifar2009sparse} is the most influential subspace clustering algorithms. This is due to its superior simplicity, accuracy, speed and robustness. SSC exploits the self-expressive property of data \cite{elhamifar2012sparse} to find the subspaces:
\begin{quote}
  {\it{each data point in a union of subspaces can be efficiently reconstructed by a combination of other points in the data}}
\end{quote}
Which gives the relation
\begin{align}
\mathbf a_i = \mathbf A \mathbf z_i.
\label{intro:ssc_model}
\end{align}
where $\mathbf z_i$ is a vector of reconstruction coefficients for $\mathbf x_i$. We can then construct a model for the entire dataset as $\mathbf A = \mathbf A \mathbf Z$. In this unrestricted case there is no unique solution for the coefficient matrix $\mathbf Z$. To resolve the dilemma SSC adopts concepts from the domain of compressed sensing, namely that among the solutions to \eqref{intro:ssc_model} 
\begin{quote}
{\it{there exists a sparse solution, $\mathbf z_i$, whose nonzero entries correspond to data points from the same subspace as $\mathbf a_i$}}.
\end{quote}
At the same time it is well known that a noiseless data point lying on $d_i$ dimensional subspace $S_i$ can be written as a linear combination of $d_i$ other noiseless points from $S_i$. Therefore since it takes at minimum $d_i$ points for the relationship to hold we can seek the sparsest representation possible i.e. a solution where the support of $z$ is smallest.

Solving such an objective is only useful when the data is known to be noise free. As previously mentioned this is extremely unlikely in practice. To overcome this SSC assumes the data model \eqref{intro:data_model} and relaxes the original constraint to yield the following objective
\begin{align}
\min_{\mathbf Z} \lambda \| \mathbf Z \|_1 + \frac{1}{2} \| \mathbf{X - X Z} \|_F^2 \quad s.t.\ \textrm{diag}(\mathbf Z) = \mathbf 0
\label{SSCRelaxedObjective}
\end{align}
where $\lambda$ is used to control the sparsity of $\mathbf Z$ and $\mathbf X$ is the observed data, opposed to $\mathbf A$, which is the latent and noise free data.

It was shown by \cite{soltanolkotabi2014robust} that correct subspace identification is guaranteed for SSC provided that data driven regularisation is used to set$\lambda_i$ (column wise splitting of $\mathbf Z$). This of course is subject to further conditions such as a minimum distance between subspaces and  sufficient sampling of points from each subspace.

Rather than compute the sparsest representation of each data point individually, Low-Rank Representation (LRR) \cite{liu2010robust} attempts to promote a more global affinity matrix by computing the lowest-rank representation of the set of data points. The LRR objective is as follows
\begin{align}
\min_{\mathbf{Z}} \;  \textrm{rank}(\mathbf{Z}), \quad
\text{s.t.} \quad \mathbf{A = AZ}.  
\end{align}
The aim of the rank penalty is to create a global grouping effect that reflects the underlying subspace structure of the data. In other words, data points belonging to the same subspace should have similar coefficient patterns.

Similar to SSC, the original objective for LRR is intractable since it involves a discrete optimisation problem. Furthermore the low-rank objective has the same problem as the $\ell_0$ objective. It is only a valid objective when the data is noiseless. Instead the authors of LRR suggest a heuristic version which is the closest convex envelope: the nuclear norm. The objective then becomes
\begin{align}
\min_{\mathbf{Z}} \;  \| \mathbf{Z} \|_*, \quad 
\text{s.t.} \quad \mathbf{X = XZ}  
\end{align}
where $\| \cdot \|_*$ is the nuclear norm of a matrix and is the sum of the singular values.

In the case where noise is present, the authors of LRR suggested a similar model to that used in SSC. However they assume that their fitting error will be only be present in a small number of columns. This results in the following objective
\begin{align}
\min_{\mathbf {E, Z}} \; \lambda \|\mathbf{E}\|_{1,2} + \|\mathbf Z\|_{*}, \quad
\text{s.t.} \quad \mathbf{X = XZ + E},
\end{align}
where $\|\mathbf{E}\|_{1,2} = \sum^n_{i=1} \| \mathbf{e_i} \|_2$ is called the $\ell_{1,2}$ norm.

\section{Collaborative Subspace Clustering and Related Work}

\begin{figure}[!t]
\centering
\subfloat[Independent Subspace Clustering]{\includegraphics[width=0.4\textwidth]{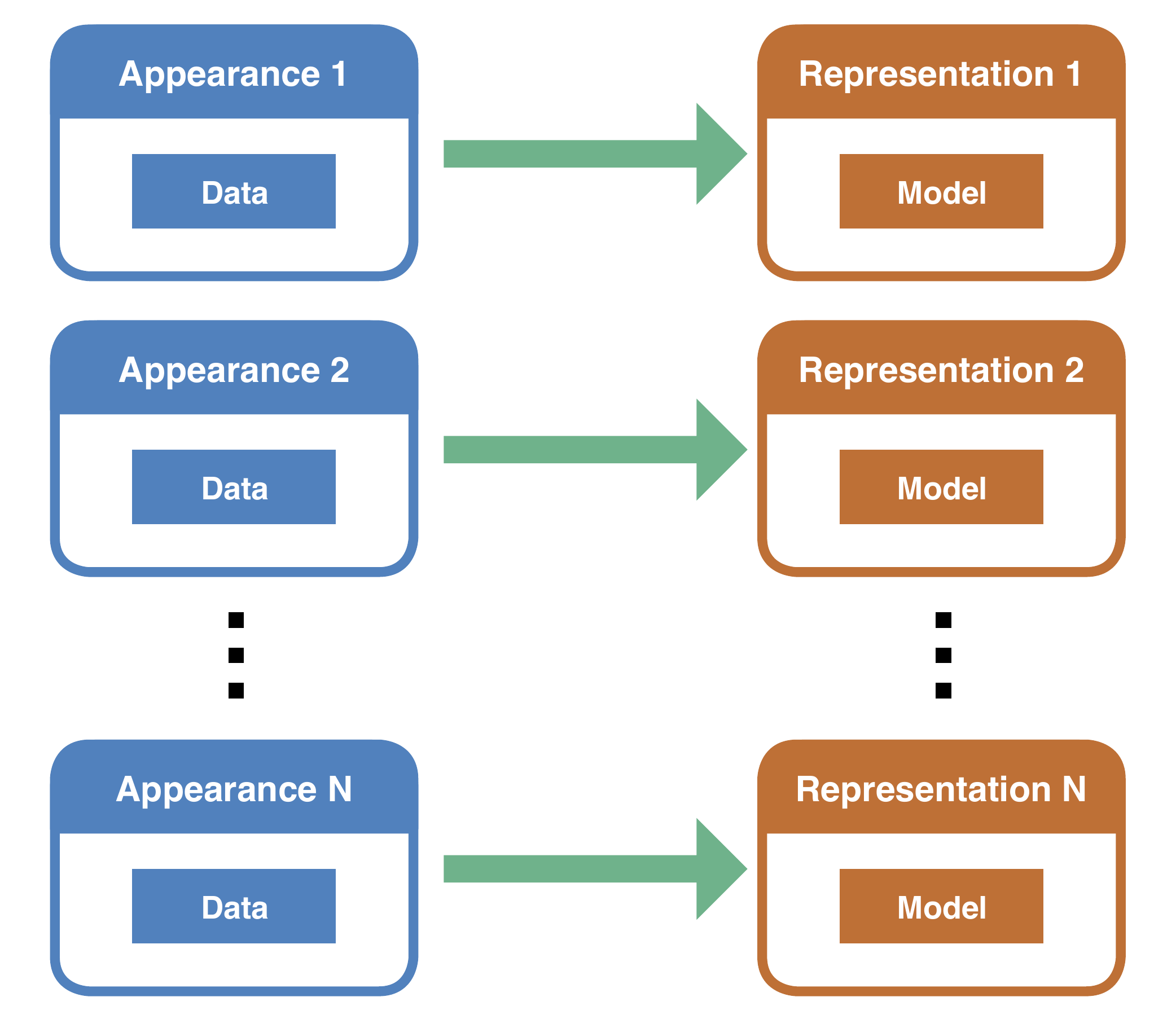}}
\subfloat[Collaborative Subspace Clustering]{\includegraphics[width=0.4\textwidth]{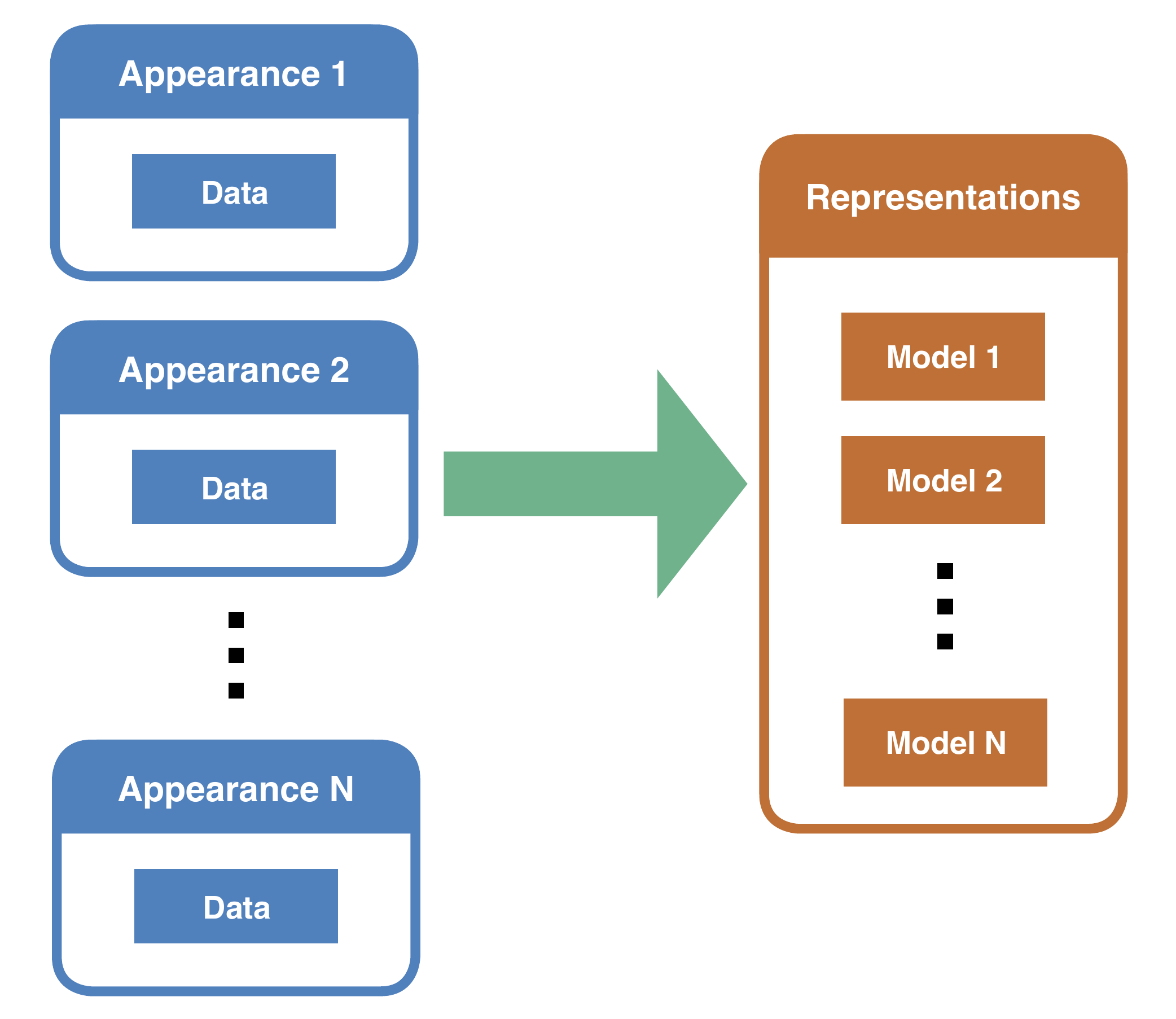}}
\caption{A comparison of independent subspace clustering and collaborative subspace clustering . In independent subspace clustering (a) each appearance and affinity is considered and learnt independently. In collaborative independent subspace clustering (b) multiple affinities are learnt jointly, utilising the cross-appearance relation.}
\label{single_multi}
\end{figure}

Prior work on subspace clustering has focused on the case with only a single observation of the phenomenon of interest. However in this paper paper we focus on learning a subspace segmentation from multiple observations. That is given a set of appearances $\{ \mathbf X_1, ..., \mathbf X_c \}$ where $c$ is the number of appearances we seek to jointly learn a single coefficient matrix $\mathbf Z$ that encodes the global coefficient structure. We illustrate this concept in Figure \ref{single_multi}.

\begin{figure}[!t]
\centering
\subfloat[]{\includegraphics[width=0.125\textwidth]{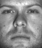}} \;
\subfloat[]{\includegraphics[width=0.125\textwidth]{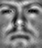}} \;
\subfloat[]{\includegraphics[width=0.125\textwidth]{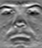}} \;
\caption{An example of tertiary appearances for face image data (Extended Yale Face Database B). (a) Original Image, (b) Laplacian of Gaussian Filtered and (c) Sobel Filtered.}
\label{FaceExample2}
\end{figure}

The motivation is two fold:
\begin{enumerate}
\item Combining information encoded in multiple observations has the potential to dramatically increase the discriminative power of the coefficient matrix since we can obtain a consensus, which should be more robust to outliers and noise. We demonstrate this in our later experiments.
\item Proliferation of sensors and reduced cost. Multi-sensor data is becoming increasingly common. For example in remote sensing, imagery is often captured from multiple sensors, each dedicated to particular frequency ranges of the electromagnetic spectrum \cite{Chen2014832, DBLP:conf/dicta/TierneyGG14a} and in human activity recognition there has been a surge in the use of RGB-Depth cameras \cite{5995316,taylor2012vitruvian}. Moreover, single observations can be turned into multiple tertiary observations. For example many image classification and recognition algorithms use many image features is used to represent the original image \cite{cheng2011multi,li2010object,le2013building}. Each of these features is actually a new observation or appearance. See Figure \ref{FaceExample2} for an example.
\end{enumerate}

A naive approach to this problem might be to simply perform spectral subspace clustering on each appearance and then fuse the coefficient matrices later. However the coefficient matrices could vary mildly, with little overlapping or shared structure between each $\mathbf Z_i$ and with vastly different magnitude of values.

The simplest solution to this problem is to simply concatenate (by row) each data matrix $\mathbf X_i$ to form a larger matrix e.g.
\[
\mathbb X = \begin{pmatrix} \mathbf X_1 \\ \mathbf X_2 \\ \dots \\ \mathbf X_c \end{pmatrix}
\]
however this has two significant drawbacks due to the heavy reliance on the Gramian matrix $\mathbf X^T \mathbf X$ in optimisation algorithms for spectral subspace clustering:
\begin{itemize}
\item  The data in each $\mathbf X_i$ may not be measuring the same quantity or at the same scale. If this is the case then some rows of each $\mathbf X_i$ will overpower others causing a loss of information. To counter this normalisation may be used.
\item The dimension of each observation ($\mathbf X_i$) may be different. Interestingly when the dimensions are different and normalisation is applied it is the data with the smallest dimension which tends to dominate and wash out the information encoded in the rest of the data.
\end{itemize}
With these two issues in mind it is clear that we must seek a different approach since concatenation is only suitable when each $\mathbf X_i$ is measuring the same quantities and is of the same dimension.

The work most closely related and which inspires our work is the so called Multi-Task Low-rank Affinity Pursuit (MLAP) \cite{cheng2011multi}. MLAP adopts the low-rank representation model for each individual observation and fuses them together by seeking sparsity consistent low-rank coefficients. The MLAP objective is the following
\begin{align}
\label{mlapobj}
\min_{\mathbf {E, Z, A}} \; \sum_i^c (\|\mathbf E_i\|_{1,2} + \lambda_i \| \mathbf Z_i \|_*) + \tau \| \mathbf A \|_{1,2} \\
\text{s.t.} \quad \mathbf X_i = \mathbf X_i \mathbf Z_i + \mathbf  E_i, \nonumber
\end{align}
where $\mathbf X_i$, $\mathbf E_i$ and $\mathbf Z_i$ are observation data, the fitting error and coefficient matrices for the $i$th observation respectively, $c$ is the number of observations, the  $\ell_{1,2}$ norm is defined as $\|\mathbf{E}\|_{1,2} = \sum^n_{i=1} \| \mathbf{e_i} \|_2$, $\mathbf A = [ \textrm{vec}(\mathbf Z_1 ), \dots, \textrm{vec}(\mathbf Z_c )  ]^T$ and $\textrm{vec}$ is the vectorisation operator such that $\textrm{vec}(\mathbf C) = \mathbf C^{m \times n} \mapsto \mathbf c^{mn}$ .


However the sparsity consistency penalty is rather unsatisfactory since it will not ensure that every $\mathbf Z_i$ is actually similar. To demonstrate this we examine the proximal $\ell_{1,2}$ thresholding step in MLAP, which can be written as
\begin{align}
\argmin_{\mathbf {A}} \; \tau \| \mathbf A \|_{1,2} + \frac{1}{2} \| \mathbf A - \mathbf V \|_F^2
\end{align}
where $\mathbf V$ is a placeholder for the variables involved in this step. The solution to this problem is given by
\begin{align}
\mathbf A_i = \begin{cases} {\displaystyle \frac{\|\mathbf v_i\|_2 - \tau}{\|\mathbf v_i\|_2}}\mathbf v_i & \textrm{if } \|\mathbf v_i\|_2 > \tau \\
0 & \textrm{otherwise}
\end{cases}
\end{align}
where $\mathbf A_i$ and $\mathbf v_i$ are columns of $\mathbf A$ and $\mathbf V$ respectively. Let
\begin{align*}
\mathbf V = \left( \begin{array}{ccc}
    0.0010  &  1.0000  &  1.0000 \\
    0.0015  &  0.1000  &  1.0100
\end{array} \right)
\end{align*}
and $\tau = 0.1$ then the solution is
\begin{align*}
\mathbf A = \left( \begin{array}{ccc}
         0  &  0.9005  &  0.9296 \\
         0  &  0.0900  &  0.9389
\end{array} \right)
\end{align*}

Here we have presented three possible scenarios in the three columns. From left to right: small magnitudes with a small difference, large and small magnitudes, and large magnitudes with a small difference. In the first and last columns where the row difference is small the row difference is further narrowed. Likewise with the middle column, however the difference is not decreased by an appreciable amount.  In other words the $\ell_{1,2}$ penalty appears to work best with values that are already close. But as previously mentioned this is unlikely to be the case as different appearances can produce vastly different coefficient matrices. In other words the $\ell_{1,2}$ penalty can fail to inductively transfer structure between the coefficient matrices. Furthermore the $\ell_{1,2}$ penalty suppresses the numeric values in $\mathbf V$. This suppression can lead to an increase in the fitting error $\mathbf E_i$ and reduce clustering accuracy.

After solving the MLAP objective \eqref{mlapobj} the authors then obtain a unified coefficient matrix $\mathbb A \in \mathbb R^{N \times N}$ by
\begin{align}
\mathbb W_{ij} = \sqrt{\sum_{k=1}^c (\mathbf Z_k)_{ij}^2} + \sqrt{\sum_{k=1}^c (\mathbf Z_k)_{ji}^2}.
\label{unify}
\end{align}
Noting that $\sqrt{\sum_k^c (\mathbf Z_k)_{ij}^2}$ is the $\ell_2$ norm of all the coefficients for element at position $i,j$.

\section{Collaborative Low-Rank Subspace Clustering}

Rather than enforce that the magnitudes of the coefficient matrices $\mathbf Z_i$ are similar, as in MLAP, we seek only that the coefficient pattern is similar. Recall that $\mathbf A = [ \textrm{vec}(\mathbf Z_1 ), \dots, \textrm{vec}(\mathbf Z_c )  ]^T$. Let us consider an ideal case where the solution coefficient pattern $\mathbf Z_i$'s are similar i.e.\ they are the scaled by a constant factor from each other. In other words they are linearly dependant and thus $\textrm{rank}(\mathbf A) = 1$. Starting from this ideal case provides our motivation. Therefore we seek to solve the following objective function
\begin{align}
\label{obj1}
&\min_{\mathbf {E, Z}} \; \sum_i \|\mathbf E_i \|_{F}^2 + \lambda_i \|  \mathbf Z_i \|_*, \\
\text{s.t.} \quad & \mathbf X_i = \mathbf X_i \mathbf Z_i + \mathbf E_i, \; \textrm{rank}(\mathbf A) = 1. \nonumber
\end{align}
Unfortunately such a strict rank condition means the objective is no longer convex, which makes optimisation difficult. We can relax the constraint into the objective function since a sufficiently large $\tau$ can simulate the $\textrm{rank}(\mathbf A) = 1$ constraint
\begin{align}
\label{obj2}
\min_{\mathbf {E, Z, A}} \; \sum_i ( \|\mathbf E_i \|_{F}^2 + \lambda_i \| \mathbf Z_i \|_*) + \tau \;\textrm{rank}(\mathbf A) \\
\text{s.t.} \quad \mathbf X_i = \mathbf X_i \mathbf Z_i + \mathbf E_i. \nonumber
\end{align}
However this objective is NP-hard. Furthermore the rank constraint may be too strict. Enforcing that each $\mathbf Z_i$ is an exact linear combination of the others may force $\mathbf E_i$ to absorb a significant amount of error. These issues lead us to our actual final objective function for LRCSC (Low-Rank Collaborative Subspace Clustering)
\begin{align}
\label{obj3}
\min_{\mathbf {E, Z, A}} \; \sum_i ( \|\mathbf E_i \|_{F}^2 + \lambda_i \| \mathbf Z_i \|_*) + \tau \| \mathbf A \|_* \\
\text{s.t.} \quad \mathbf X_i = \mathbf X_i \mathbf  Z_i + \mathbf E_i \nonumber
\end{align}
We now have a feasible objective function that can simulate the rank constraint. Just as $\tau$ can simulate the rank one constraint in \eqref{obj2}, $\tau$ in \eqref{obj3} can also simulate this constraint if it is sufficiently large \cite{cabral2013unifying}. If a unified coefficient matrix is required we take the approach described in \eqref{unify}.

We outline the full Collaborative Low-Rank Subspace Clustering (cLRSC) algorithm in Algorithm \eqref{alg_final}.  Compared with MLAP, cLRSC will  enforce coefficients in each affinity matrix to have a similar pattern rather than similar magnitudes. We later show through experimental evaluation that cLRSC outperforms MLAP consistently.

\begin{algorithm}[!t]
\caption{cLRSC}
\label{alg_final}
\begin{algorithmic}[1]

\REQUIRE $\{\mathbf X^{D \times N}\}$ - observed data, $c$ - number of subspaces

\STATE Obtain coefficients $\mathbf Z_i$ by solving \eqref{obj3}

\STATE Form the similarity graph $\mathbb W_{ij} = \sqrt{\sum_{k=1}^c (\mathbf Z_k)_{ij}^2} + \sqrt{\sum_{k=1}^c (\mathbf Z_k)_{ji}^2}.$

\STATE Apply N-Cut to $\mathbf W$ to partition the data into $c$ subspaces

\RETURN Subspace label indices for $\{S_i\}^c_{i=1}$

\end{algorithmic}
\end{algorithm}

\subsection{Optimisation}

To solve \eqref{obj3} we take the path provided by LADMPSAP (Linearized Alternating Direction Method with Parallel Spliting and Adaptive Penalty) \cite{DBLP:conf/acml/LiuLS13}. An overview of the entire algorithm can be found in Algorithm \ref{collabalg}. Here we expand on solving the primary variables. First introduce an auxiliary constraint to explicitly enforce the relationship between $\mathbf Z_i$ and $\mathbf A$
\begin{align}
\min_{\mathbf {E, Z, A}} \; \sum_{i=1}^c (\| \mathbf E_i \|_{F}^2 + \lambda_i \| \mathbf Z_i \|_*) + \tau \| \mathbf A \|_* \\
\text{s.t.} \quad \mathbf X_i = \mathbf X_i \mathbf Z_i + \mathbf E_i \nonumber \\
\textrm{vec}(\mathbf Z_i)^T = \mathbf A_{(i, :)} \nonumber 
\end{align}
where $\mathbf A_{(i, :)}$ is the $i$th row of $\mathbf A$.
Next we relax the constraints and form the Augmented Lagrangian
\begin{align}
&\min_{\mathbf E_i, \mathbf Z_i, \mathbf A} \; \sum_{i=1}^c ( \| \mathbf E_i \|_F^2 + \lambda_i \|  \mathbf Z_i \|_* ) + \tau \| \mathbf A \|_* \nonumber \\
&+ \sum_{i=1}^c ( \langle \mathbf Y_i, \mathbf X_i - \mathbf X_i \mathbf Z_i - \mathbf E_i \rangle  + \frac{\mu}{2} \| \mathbf X_i - \mathbf  X_i \mathbf Z_i - \mathbf E_i \|_F^2\nonumber \\
&+  \langle \mathbf w_i, \textrm{vec}(\mathbf Z_i)^T - \mathbf A_{(i, :)} \rangle + \frac{\mu}{2} \| \textrm{vec}(\mathbf Z_i)^T - \mathbf A_{(i, :)} \|_F^2 )
\end{align} 

Each $\mathbf Z_i$ can be updated in parallel by 
\begin{align*}
\min_{\mathbf Z_i} \;   \sum_{i=1}^c ( \lambda_i \|  \mathbf Z_i \|_*  
+ \frac{\mu}{2} \| \mathbf X_i - \mathbf X_i \mathbf Z_i - \mathbf E_i + \frac{1}{\mu} \mathbf Y_i \|_F^2\\
+ \frac{\mu}{2} \| \textrm{vec}(\mathbf Z_i)^T - \mathbf A_{(i, :)} + \frac{1}{\mu} \mathbf w_i \|_F^2 )
\end{align*}
To find a closed form solution we linearise the second and third terms
\begin{align*}
\min_{\mathbf Z_i} \;   \sum_{i=1}^c ( \lambda_i \| \mathbf Z_i \|_*  + \frac{\rho}{2} \| \mathbf Z_i - (\mathbf Z_i^t - \frac{1}{\rho} \partial G(\mathbf Z_i^t)) \|_F^2 )
\end{align*}
where $G = \frac{\mu}{2} \| \mathbf X_i - \mathbf X_i \mathbf Z_i - \mathbf E_i + \frac{1}{\mu} \mathbf Y_i \|_F^2 + \frac{\mu}{2} \| \textrm{vec}(\mathbf Z_i)^T - \mathbf A_{(i, :)} + \frac{1}{\mu} \mathbf w_i \|_F^2$ and $\partial G = - \mu \mathbf X^T ( \mathbf X_i - \mathbf X_i \mathbf Z_i - \mathbf E_i + \frac{1}{\mu} \mathbf Y_i) + \mu (\textrm{vec}(\mathbf Z_i)^T - \mathbf A_{(i, :)} + \frac{1}{\mu} \mathbf w_i)$. Note that the second half of $\partial G$ needs to be reshaped into the same size as $\mathbf Z_i$. The solution to this problem is given by
\begin{align}
\label{SolutionZ}
&\mathbf Z_i = \mathbf U S_{\frac{\lambda_i}{\rho}}(\mathbf \Sigma) \mathbf V^T,\\
S_{\beta}(\mathbf \Sigma) &= \text{diag}(\{\text{max}(\sigma_i - \beta, 0)\}), \nonumber
\end{align}
where $\mathbf{ U \Sigma \mathbf V}^T$ is the Singular Value Decomposition (SVD) of $\mathbf Z_i^t - \frac{1}{\rho} \partial G(\mathbf Z_i^t)$ and $\sigma_i$ are the singular values, see \cite{parikh2013proximal}\cite{cai2010singular} for details.

Each $\mathbf E_i$ can be updated in parallel by
\begin{align*}
\min_{\mathbf E_i} \; \sum_{i=1}^c ( \| \mathbf E_i \|_F^2 + \frac{\mu}{2} \| \mathbf X_i - \mathbf X_i \mathbf Z_i - \mathbf E_i + \frac{1}{\mu} \mathbf Y_i \|_F^2 )
\end{align*}
\begin{align}
\label{SolutionE}
\mathbf E_i = \frac{\mathbf X_i - \mathbf X_i \mathbf Z_i - \frac{1}{\mu} \mathbf Y_i}{\frac{1}{\mu^k} + 1}
\end{align}

$\mathbf A$ can be updated in parallel by
\[
\min_{\mathbf {A}} \; \tau \| \mathbf A \|_* +  \sum_{i=1}^c ( \frac{\mu}{2} \| \textrm{vec}(\mathbf Z_i)^T - \mathbf A_{(i, :)} + \frac{1}{\mu} \mathbf w_i \|_F^2 )
\]
which we rearrange to
\[
\min_{\mathbf {A}} \; \frac{\tau}{\mu} \| \mathbf A \|_* +  \frac{1}{2} \| \mathbf A - \mathbf B \|_F^2
\]
where $\mathbf B_{(i, :)} = \textrm{vec}(\mathbf Z_i)^T + \frac{1}{\mu} \mathbf w_i$ and the final solution is given by
\begin{align}
\label{SolutionA}
&\mathbf A = \mathbf U S_{\frac{\tau}{\mu}}(\mathbf \Sigma) \mathbf V^T.
\end{align}
where $\mathbf{ U \Sigma \mathbf V}^T$ is the SVD of $\mathbf B$.

\begin{algorithm}
\caption{{\bf Solving \eqref{obj3} by LADMPSAP}}
\label{collabalg}
\begin{algorithmic}[1]

\REQUIRE $\{ \mathbf X_i, ... \mathbf X_c \}$, $\{ \lambda_i, ... \lambda_c \}$, $\tau$, $\mu$, $\mu^{\text{max}} >> \mu$, $\rho > \| \mathbf X \|^2$ and $\epsilon_1, \epsilon_2 > 0$.

\STATE Initialise $\{ \mathbf Z_i, ... \mathbf Z_c \} = \mathbf 0$, $\{ \mathbf E_i, ... \mathbf E_c \} = \mathbf 0$, $\mathbf A = \mathbf 0$, $\{ \mathbf Y_i, ... \mathbf Y_c \} = \mathbf 0$, $\{ \mathbf w_i, ... \mathbf w_c \} = \mathbf 0$

\WHILE{not converged}

\STATE Update each $\mathbf Z_i^{k+1}$ by \eqref{SolutionZ}
\STATE Update each $\mathbf E_i^{k+1}$ by \eqref{SolutionE}
\STATE Update $\mathbf A^{k+1}$ by \eqref{SolutionA}

\STATE Set $m$
\begin{align*}
m = \frac{\mu^k \sqrt{\rho}}{\| \mathbf X \|_F}\textrm{max} \big\{ &\| \mathbf A^{k+1} -  \mathbf A^{k}  \|_F,\\
& \| \mathbf Z_i^{k+1} - \mathbf Z_i^{k}  \|_F, \|  \mathbf E_i^{k+1} - \mathbf E_i^{k} \|,\\
&\dots,\\
&\| \mathbf Z_c^{k+1} - \mathbf Z_c^{k}  \|_F, \|  \mathbf E_c^{k+1} - \mathbf E_c^{k} \|)\big\}
\end{align*}

\STATE Check stopping criteria for each $i$ to $c$
\begin{align*}
&\frac{\|\mathbf X \mathbf Z_i^{k+1} - \mathbf X + \mathbf E_i^{k+1} \|_F}{\| \mathbf X_i \|_F} < \epsilon_1;
 \\
&\frac{\|\mathbf A_{(i, :)}^{k+1} - \textrm{vec}(\mathbf Z_i^{k+1})^T\|_F}{\| \mathbf X_i \|_F} < \epsilon_1;
\\
&m < \epsilon_2
\end{align*}

\STATE Update each $\mathbf Y_i$\\
$\mathbf Y_i^{k+1} = \mathbf Y_i^k + \mu ( \mathbf X_i - \mathbf X_i \mathbf Z_i - \mathbf E_i )$

\STATE Update each $\mathbf w_i$\\
$\mathbf w_i^{k+1} = \mathbf w_i^k + \mu ( \textrm{vec}(\mathbf Z_i)^T - \mathbf A_{(i, :)})$

\STATE Update $\gamma$
\[
\gamma_1 =
\begin{cases}
\gamma^0 & \text{if} \;\; m < \epsilon_2 \\
1 & \text{otherwise,}
\end{cases}
\]

\STATE $\mu^{k+1} = \textrm{min}( \mu_{\text{max}}, \gamma \mu^k)$

\ENDWHILE
%
%
\end{algorithmic}
\end{algorithm}

\section{Experimental Evaluation}

In this section we evaluate cLRSC for subspace clustering. First we perform a synthetic experiment with randomly generated appearance data generated. Next we perform a semi-synthetic experiment generated from a library of real data. Finally we perform an experiment on real world data in the form of face images. Parameters were fixed for each experiment and tuned for best performance.

In an effort to maximise transparency and repeatability, all MATLAB code and data used for these experiments and those in Section \ref{Section:experiments} can be found online at \url{https://github.com/sjtrny/collab_lrsc}.

To compare clustering accuracy we use the subspace clustering accuracy (SCA) metric \cite{elhamifar2012sparse}, which is defined as
\begin{align}
\text{SCA} = 100 - \frac{\text{num. of misclassified points $\times 100$}}{\text{total num. of points}}
\end{align}
Essentially it is a percentage of possible accuracy i.e. $100$ is perfect clustering while $0$ has no matched clusters. However due to ambiguity such a low score is not possible. In cases where we inject extra noise we report the level of noise using Peak Signal-to-Noise Ratio (PSNR) defined as
\begin{align}
\text{PSNR} = 10 \log_{10} \left( \frac{s^2}{\frac{1}{m n} \sum_i^{m} \sum_j^{n} ( A_{ij} - X_{ij})^2} \right)
\label{PSNR}
\end{align}
where $\mathbf{X = A + N}$ is the noisy data and $s$ is the maximum possible value of an element of $\mathbf A$. Decreasing values of PSNR indicate increasing amounts of noise.

\subsection{Synthetic Data}

In this section we compare clustering accuracy from a randomly generated dataset. Similar to \cite{liu2010robust} we construct $5$ subspaces $\{S_i\}^5_{i=1}$ whose bases $\{\mathbf U_i\}^5_{i=1}$ are computed by $\mathbf U_{i+1} = \mathbf T \mathbf U_i, 1 \leq i \leq 4$, where $\mathbf T$ is a random rotation matrix and $\mathbf U_1$ is a random orthonormal basis of dimension $100 \times 4$. In other words each basis is a random rotation away from the previous basis and the dimension of each subspace is $4$. $20$ data points are sampled from each subspace by $\mathbf X_i = \mathbf U_i \mathbf Q_i$ with $\mathbf Q_i$ being a $4 \times 20$ random iid $\mathcal{N}(0,1)$ matrix. The data is concatenated $\mathbb X = [\mathbf X_1, \mathbf X_2,\dots, \mathbf X_{5}]$. This process is repeated again yielding two appearances $\{ \mathbb X_1 , \mathbb X_2 \}$.

\begin{figure*}[!ht]
\centering
\subfloat[Mean SCA]{\includegraphics[width=0.45\textwidth]{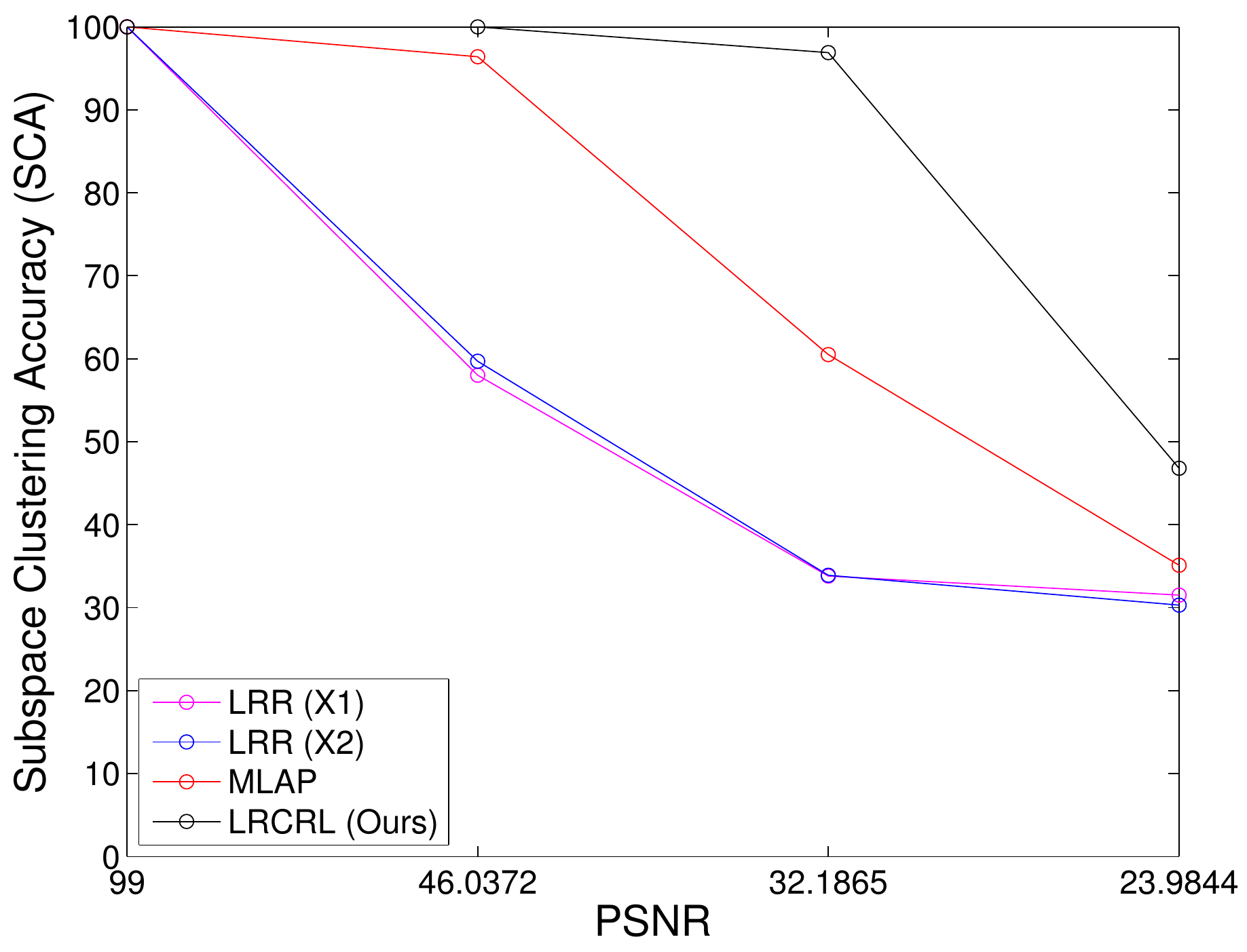}}
\subfloat[Median SCA]{\includegraphics[width=0.45\textwidth]{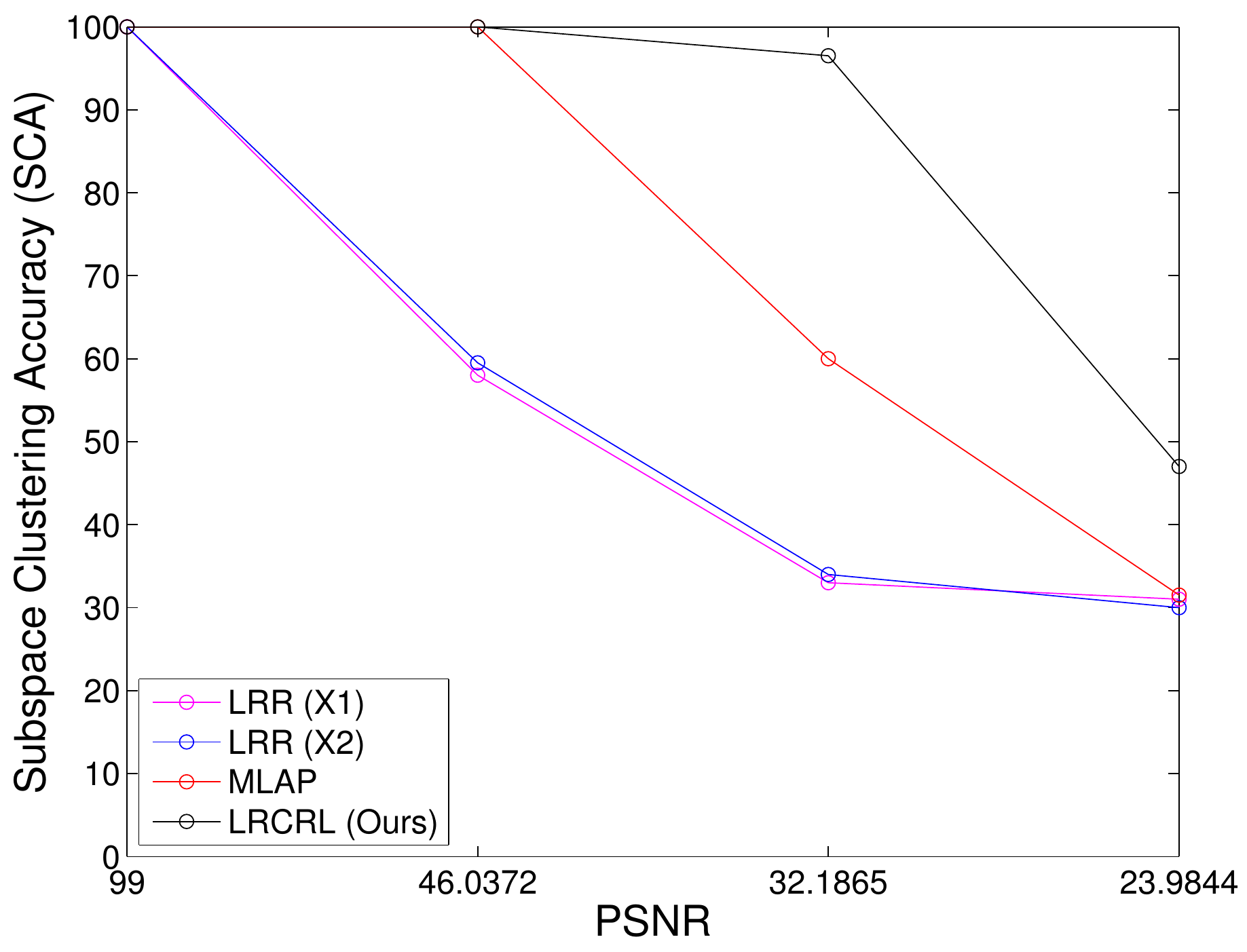}}\\
\subfloat[Maximum SCA]{\includegraphics[width=0.45\textwidth]{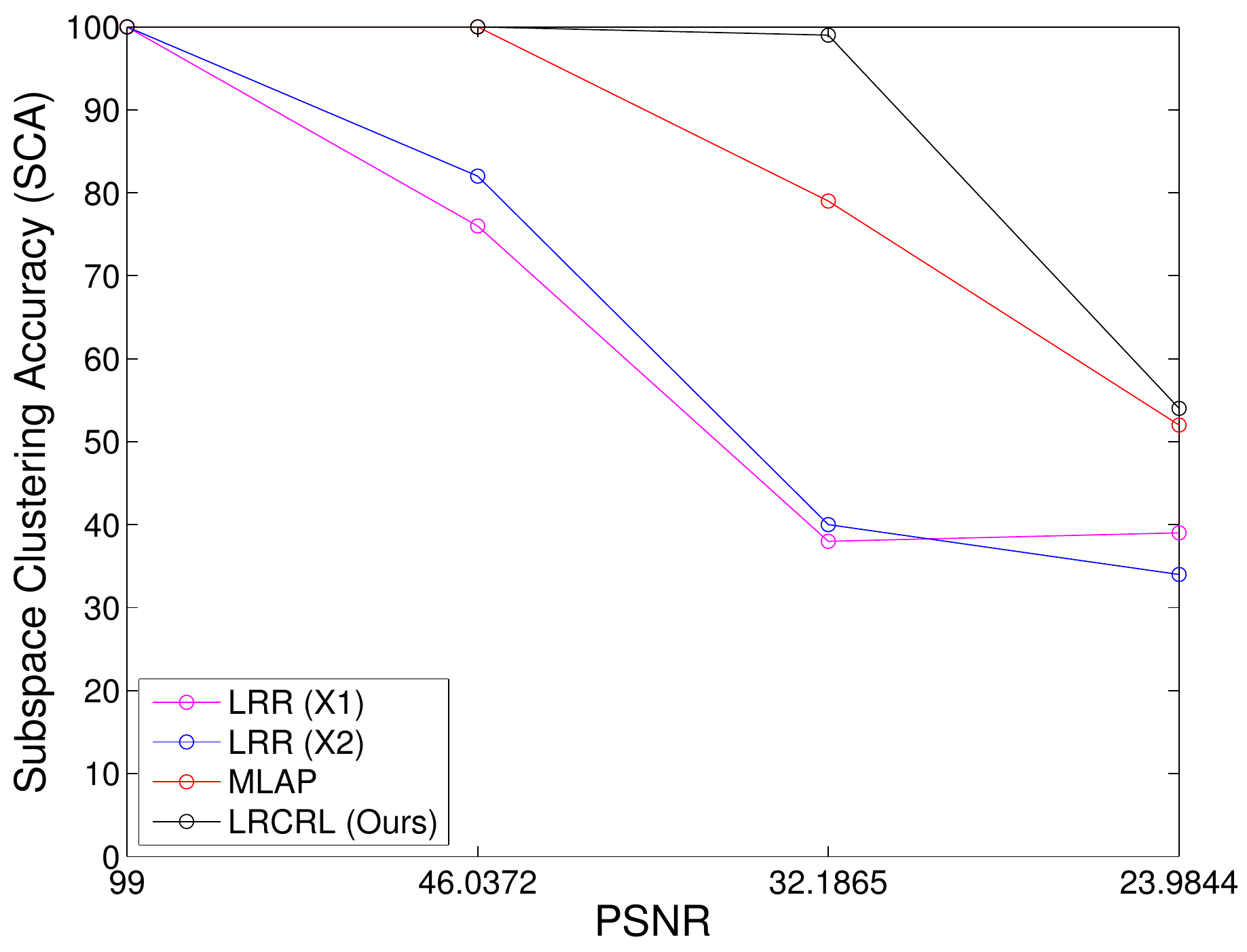}}
\subfloat[Minimum SCA]{\includegraphics[width=0.45\textwidth]{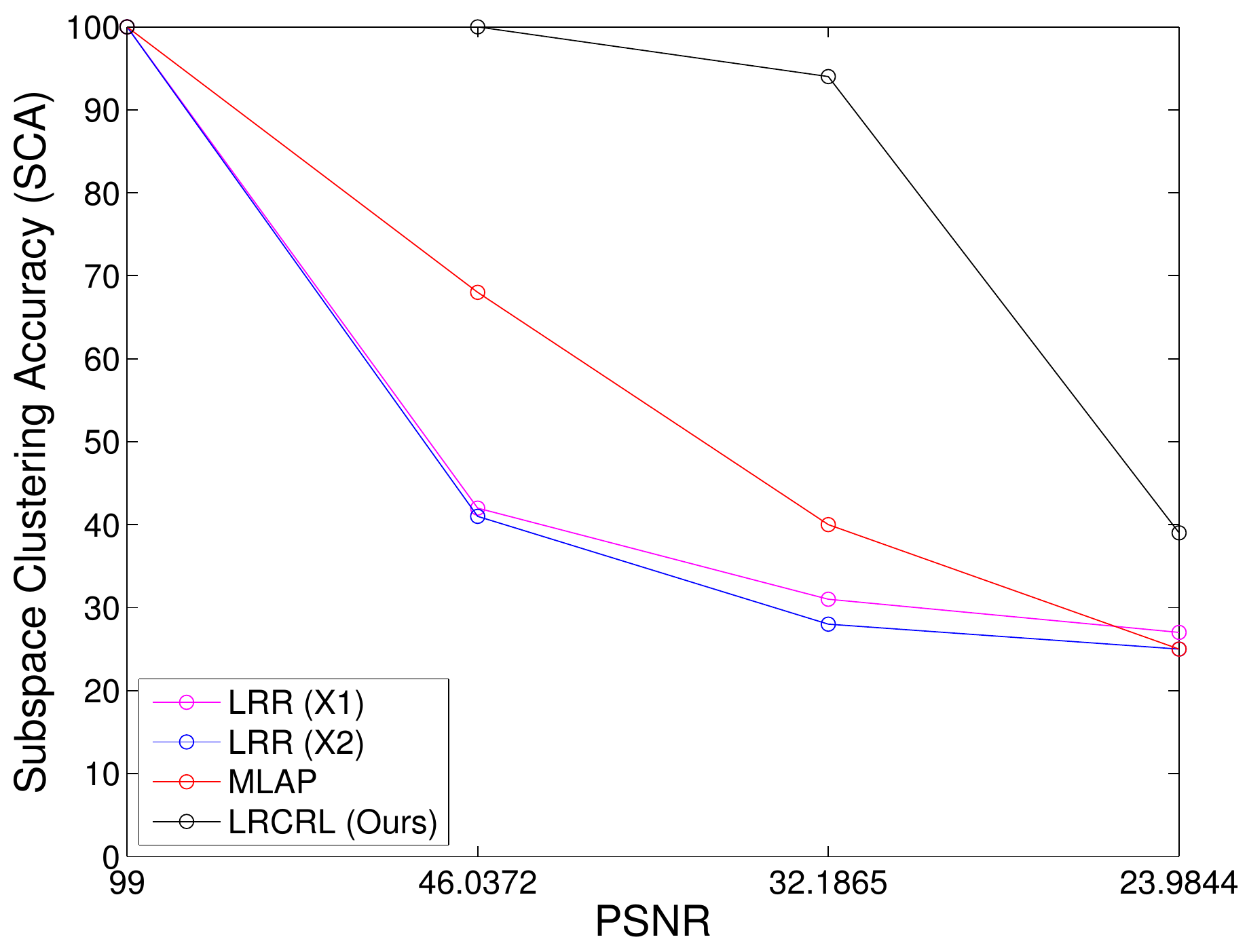}}
\caption{Results for the synthetic data segmentation experiment with various magnitudes of Gaussian noise. cLRSC (ours) outperforms both MLAP and LRR at all noise levels.}
\label{Plot:SyntheticStats}
\end{figure*}

We perform subspace clustering with LRR on the individual appearances, MLAP and cLRSC on both observations. Normalized cuts \cite{shi2000normalized} is used for final segmentation. To obtain accurate results we repeat this experiment $50$ times. Further to test robustness we repeat again for various levels of corruption by Gaussian noise. Results are reported in Figure \ref{Plot:SyntheticStats}. cLRSC (ours) outperforms both MLAP and LRR at all noise levels. This demonstrates our earlier hypothesis that collaborative learning of can increase the discriminative power  of coefficient matrices. Further cLRSC shows robustness even in cases of extreme noise. For example Figure \ref{Plot:SyntheticStats} (d) the minimum SCA at PSNR $32$ is $94\%$. In contrast minimum SCA of MLAP for this case is $40\%$.

\begin{figure}[!t]
\centering
\includegraphics[width=0.4\textwidth]{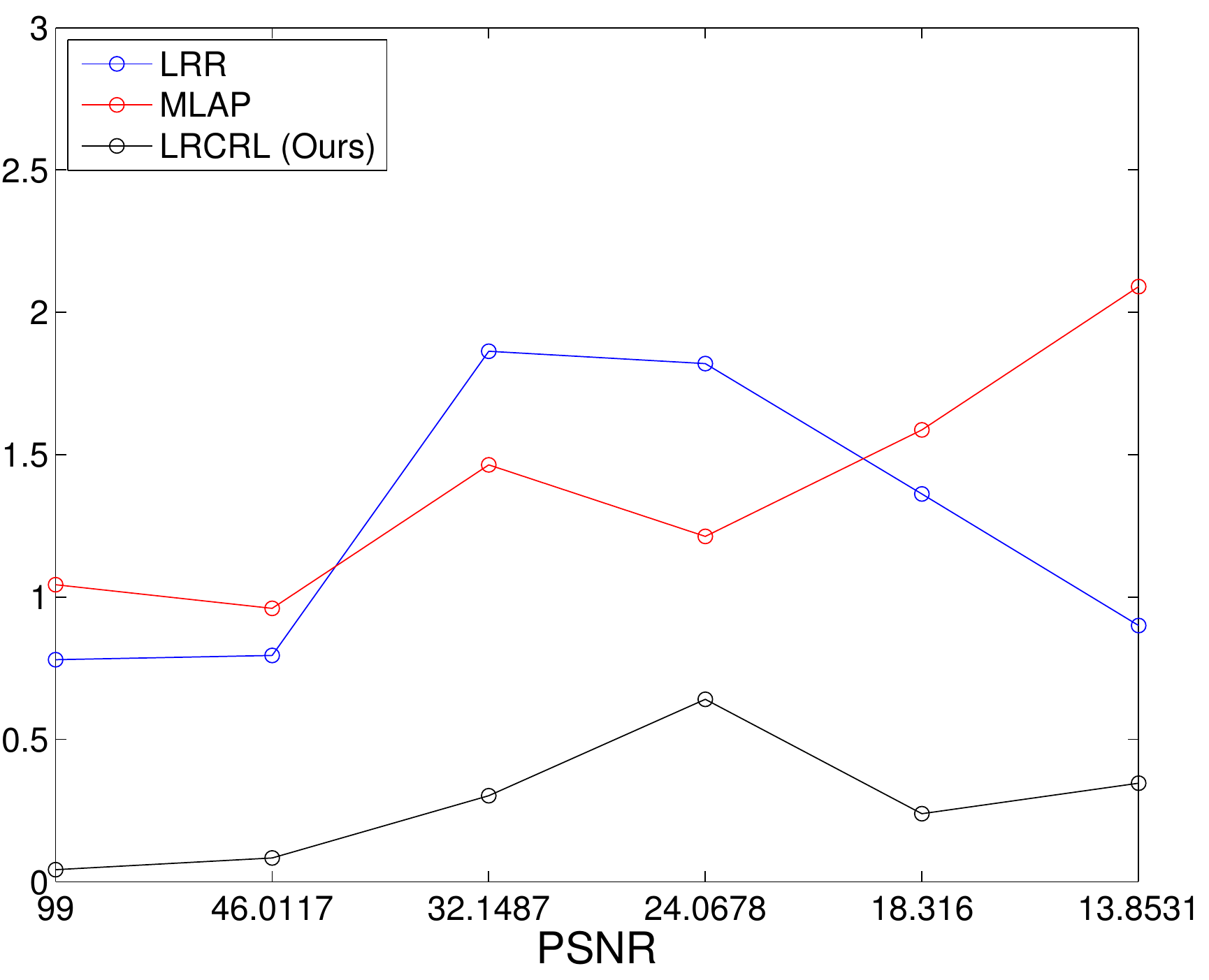}
\caption{A comparison of mean differences between each $\mathbf Z_i$ for the semi-synthetic dataset.}
\label{Plot:TIRDiff}
\end{figure} 

\subsection{Semi-Synthetic Data}

We assemble semi-synthetic data from a library of pure infrared hyper spectral mineral data. We randomly take $5$ pure spectra samples from the library such that $\mathbf A_i = [\mathbf a_1, \mathbf a_2, \dots, \mathbf a_5] \in \mathbb{R}^{321 \times 5}$. Next we combine these samples into a single synthetic sample using uniform random weights $\mathbf w_i \in \mathbb{R}^{5}$ such that $\mathbf x_i \in \mathbb{R}^{321} = \mathbf A_i \mathbf w_i$. We then repeat $\mathbf x_i$ 10 times column-wise giving $\mathbf X_i \in \mathbb{R}^{321 \times 10}$. We repeat this process $5$ times and combine all $\mathbf X_i$ to create our artificial data $\mathbb X \in \mathbb{R}^{321 \times 50} = [\mathbf X_1, \mathbf X_2,\dots, \mathbf X_{5}]$. The entire process is repeated again yielding two observations of the same phenomenon $\{ \mathbb X_1 , \mathbb X_2 \}$.

\begin{figure*}[!ht]
\centering
\subfloat[Mean SCA]{\includegraphics[width=0.45\textwidth]{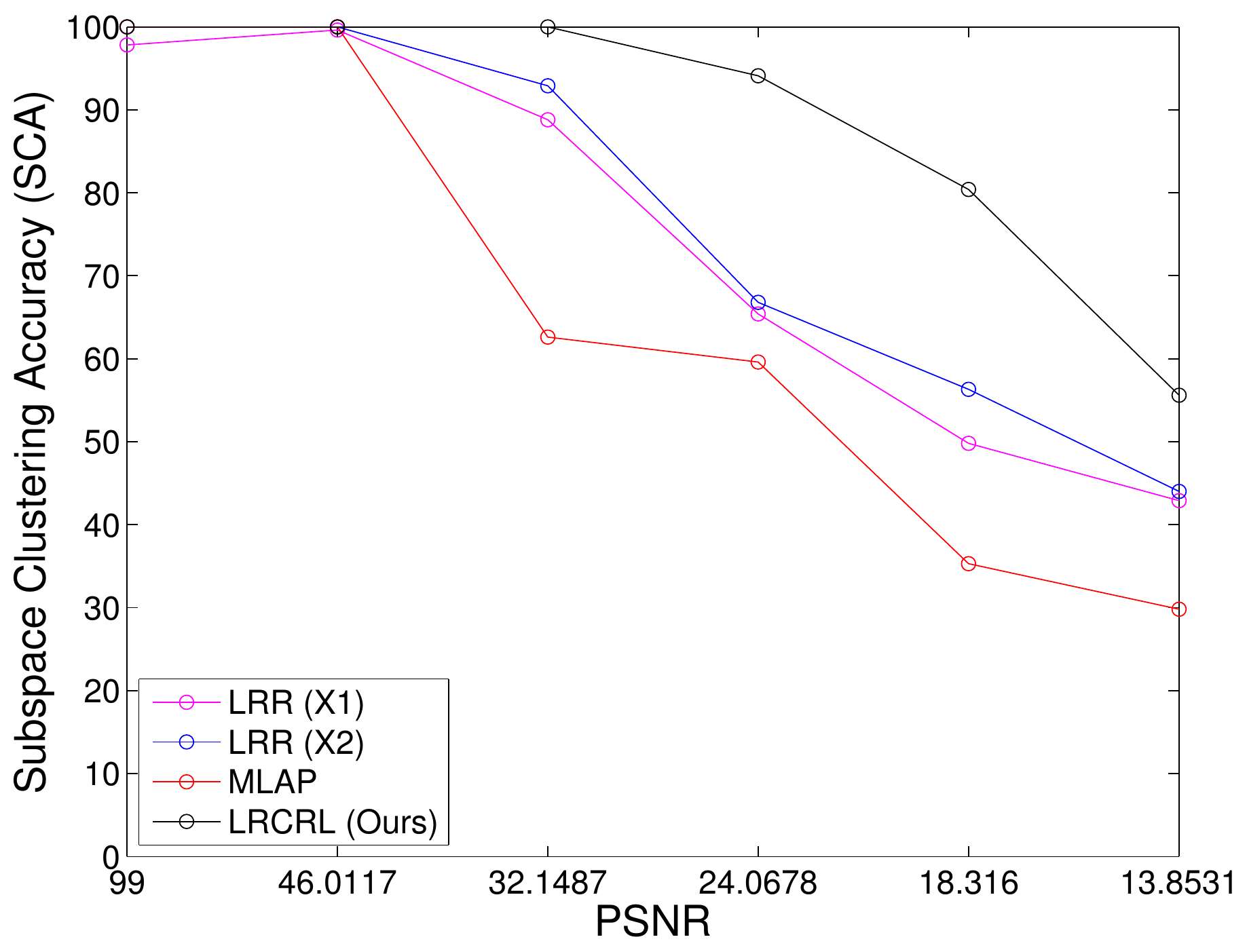}}
\subfloat[Median SCA]{\includegraphics[width=0.45\textwidth]{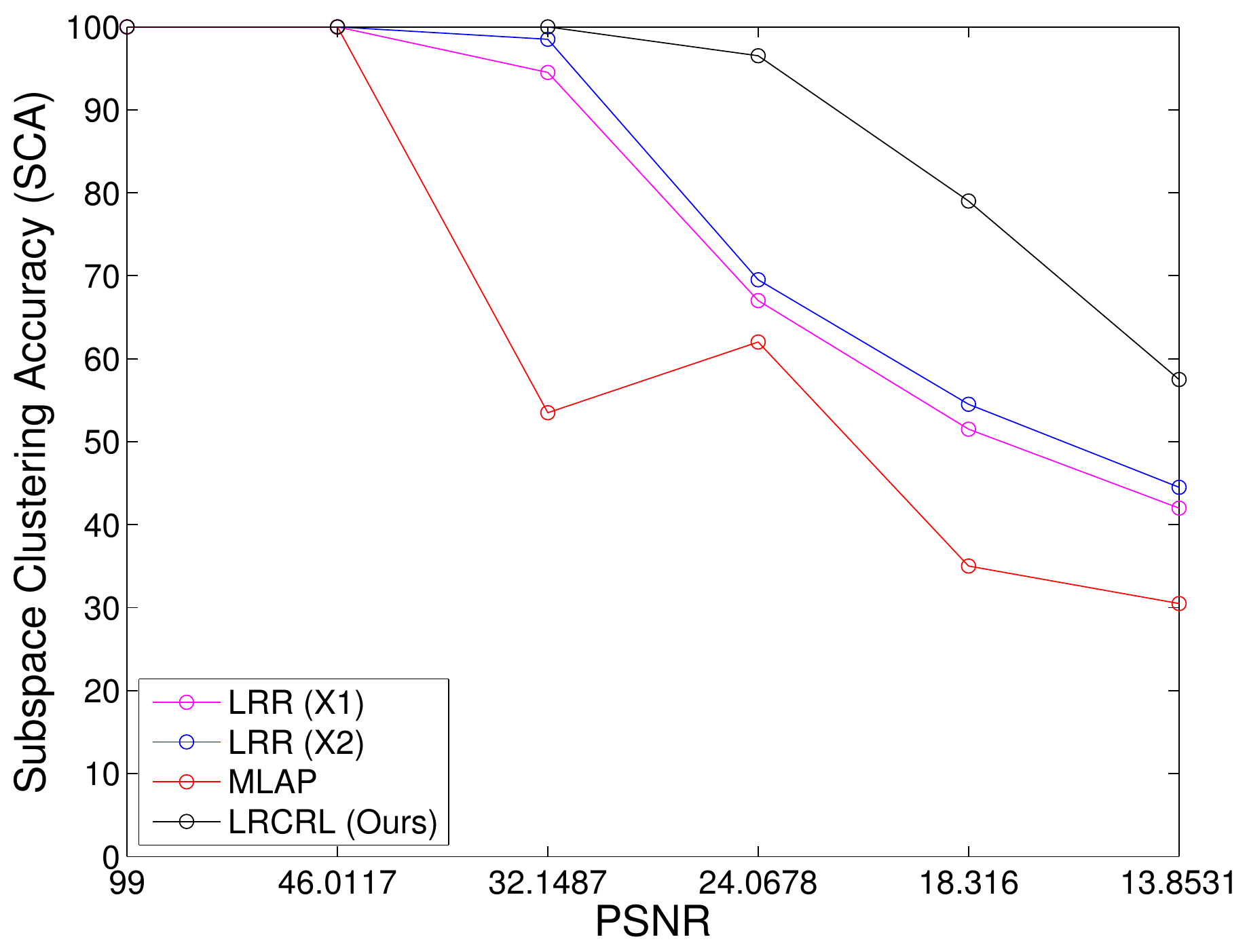}}\\
\subfloat[Maximum SCA]{\includegraphics[width=0.45\textwidth]{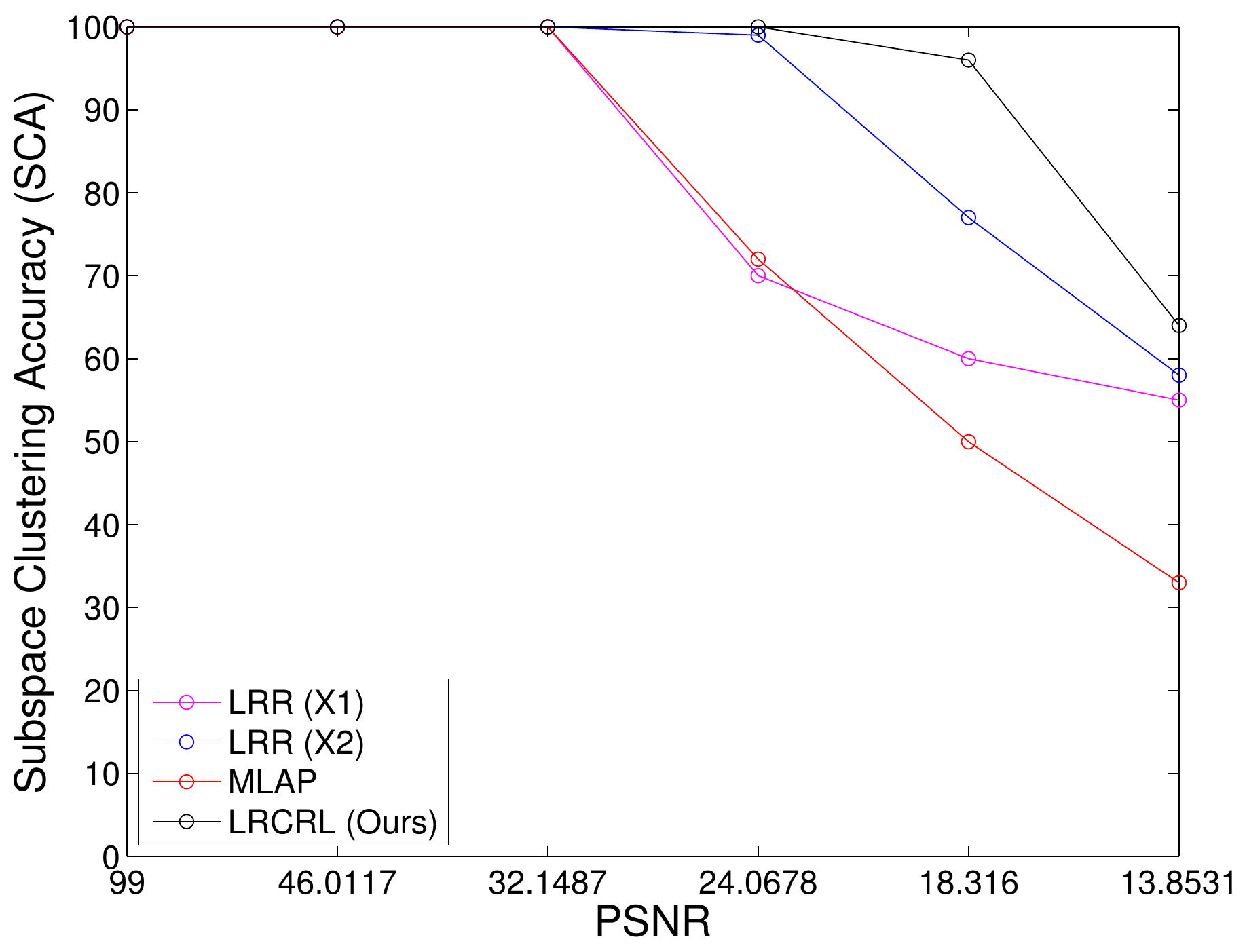}}
\subfloat[Minimum SCA]{\includegraphics[width=0.45\textwidth]{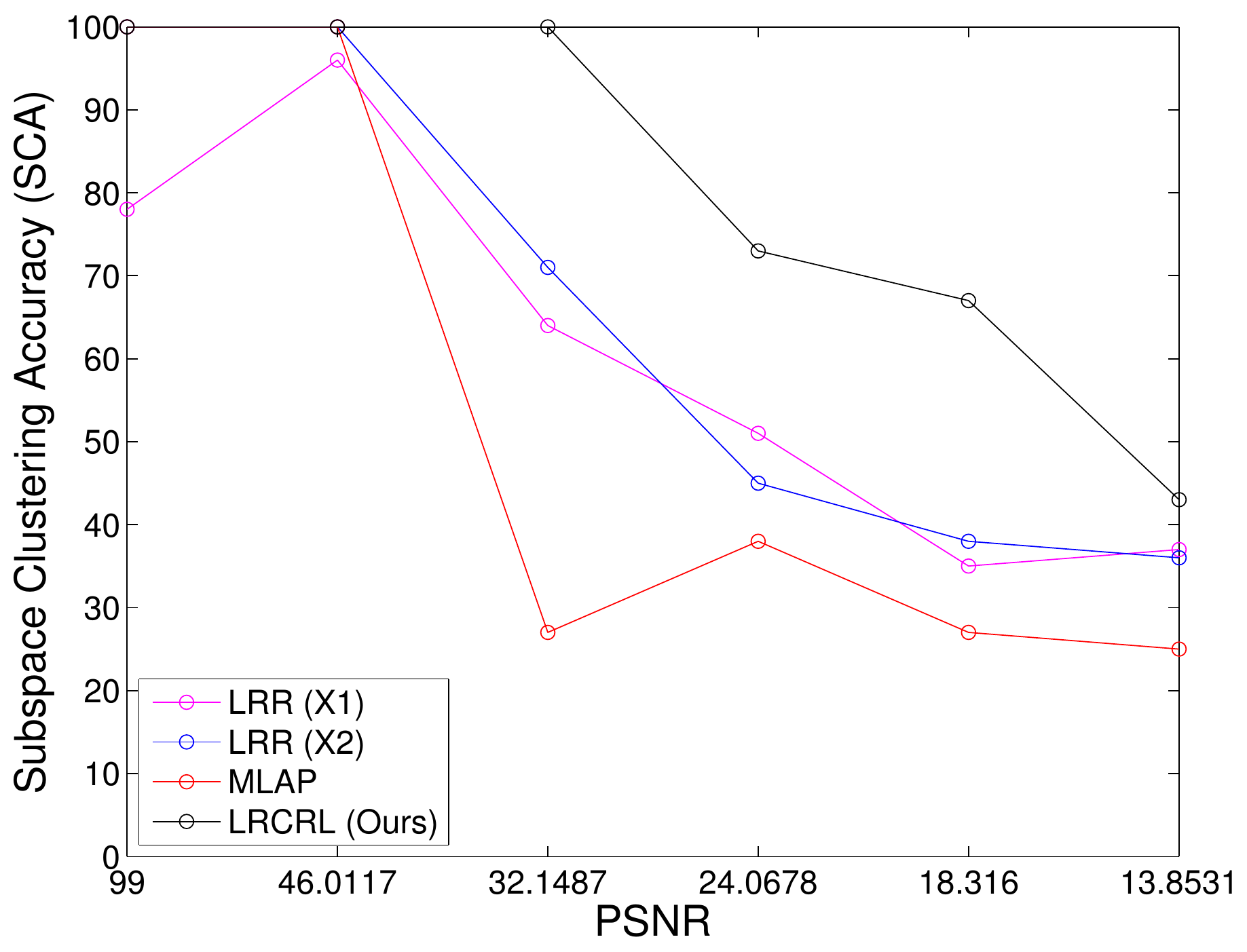}}
\caption{Results for the semi-synthetic data segmentation experiment with various magnitudes of Gaussian noise. cLRSC (ours) outperforms both MLAP and LRR at all noise levels.}
\label{Plot:TIRStats}
\end{figure*}

We perform subspace clustering with LRR on the individual appearances, MLAP and cLRSC on both observations. Normalized cuts \cite{shi2000normalized} is used for final segmentation. To obtain accurate results we repeat this experiment $50$ times. Further to test robustness we repeat again for various levels of corruption by Gaussian noise. Results are reported in Figure \ref{Plot:TIRStats}. Like the previous synthetic experiment cLRSC (ours) outperforms both MLAP and LRR at all noise levels and is far more robust at extreme noise levels.

In Figure \ref{Plot:TIRDiff} we compare how close the coefficient matrices $\mathbf Z_1$ and $\mathbf Z_2$ are for cLRSC, MLAP and LRR. We measure the mean difference through the Frobenius norm i.e.\ $\frac{\sum_1^{50}\| \mathbf Z_1^i - \mathbf Z^i_2 \|_F}{50}$. From the results we can see that the coefficient matrices are much closer under cLRSC than MLAP and LRR. Note that in most cases the difference between MLAP coefficient matrices is actually larger than those from LRR on individual appearance data. This highlights our earlier point that MLAP cannot jointly learn coefficient matrices.

\subsection{Real Data}

The aim of this experiment is to segment unique human subjects from a set of face images. We draw our data from the Extended Yale Face Database B \cite{georghiades2001few}. The dataset consists of approximately of 38 subjects under varying illumination (see Figure \ref{FaceExample2}). The test consisted of randomly selecting 5 subjects and their corresponding exemplar image (minimum shadows and corruption). These images were then resampled to $42 \times 48$ pixels and then repeated 20 times column wise and finally concatenated together to yield $\mathbb X_0 \in \mathbb{R}^{2016 \times 100} = [\mathbf X_1, \mathbf X_2,\dots, \mathbf X_{5}]$ where $\mathbf X_i \in \mathbb{R}^{2016 \times 20}$.

\begin{figure*}[]
\centering
\subfloat[Mean SCA]{\includegraphics[width=0.45\textwidth]{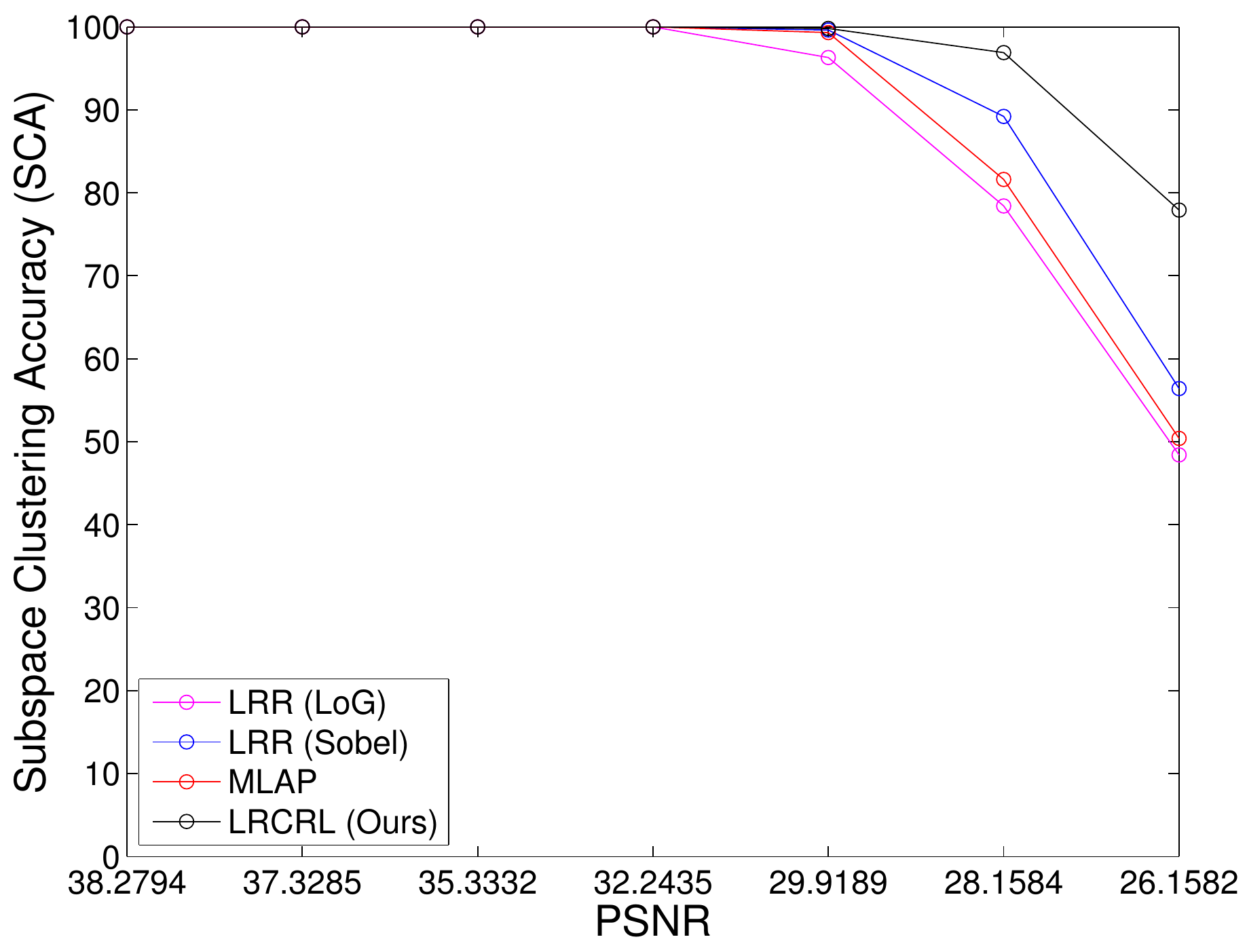}}
\subfloat[Median SCA]{\includegraphics[width=0.45\textwidth]{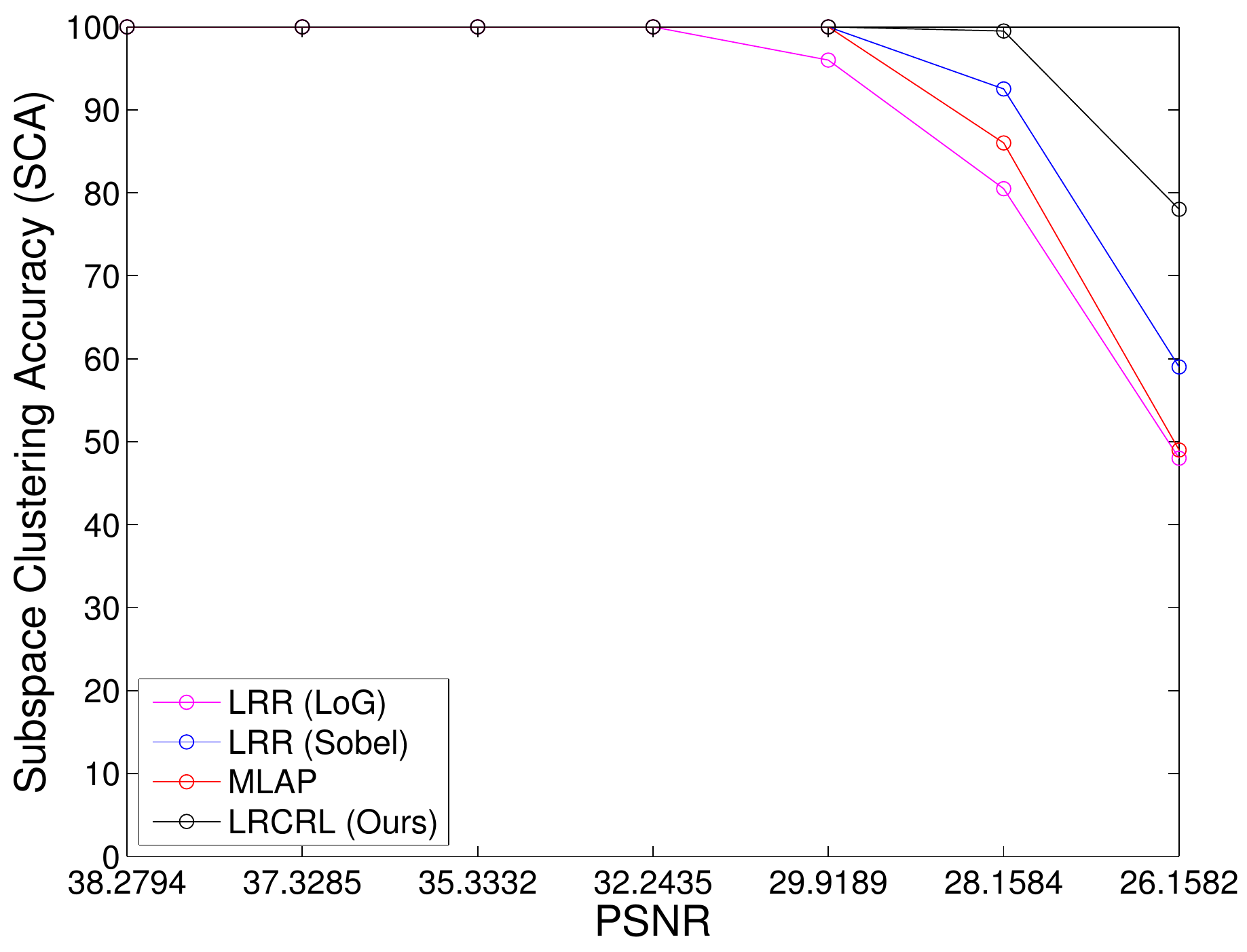}}\\
\subfloat[Maximum SCA]{\includegraphics[width=0.45\textwidth]{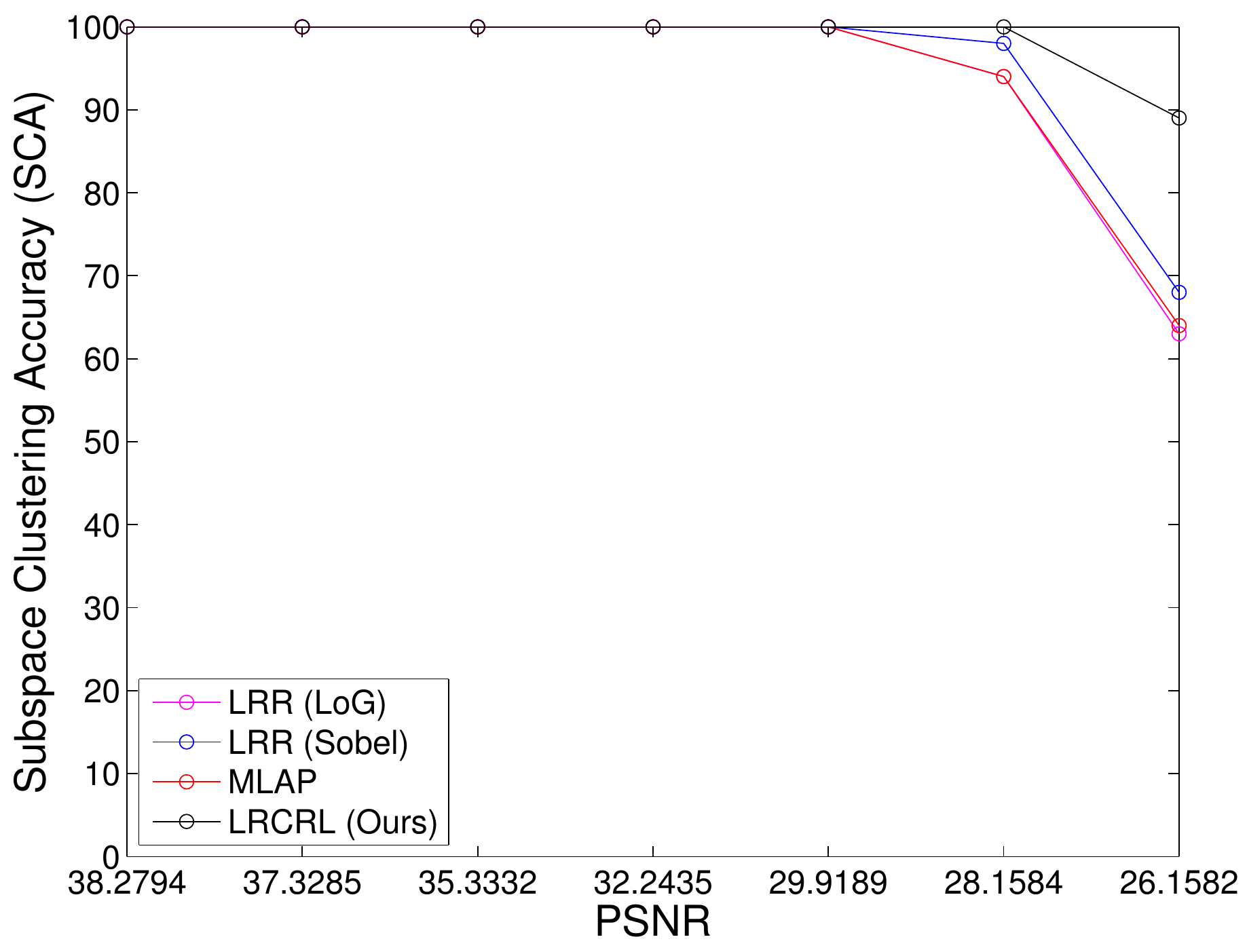}}
\subfloat[Minimum SCA]{\includegraphics[width=0.45\textwidth]{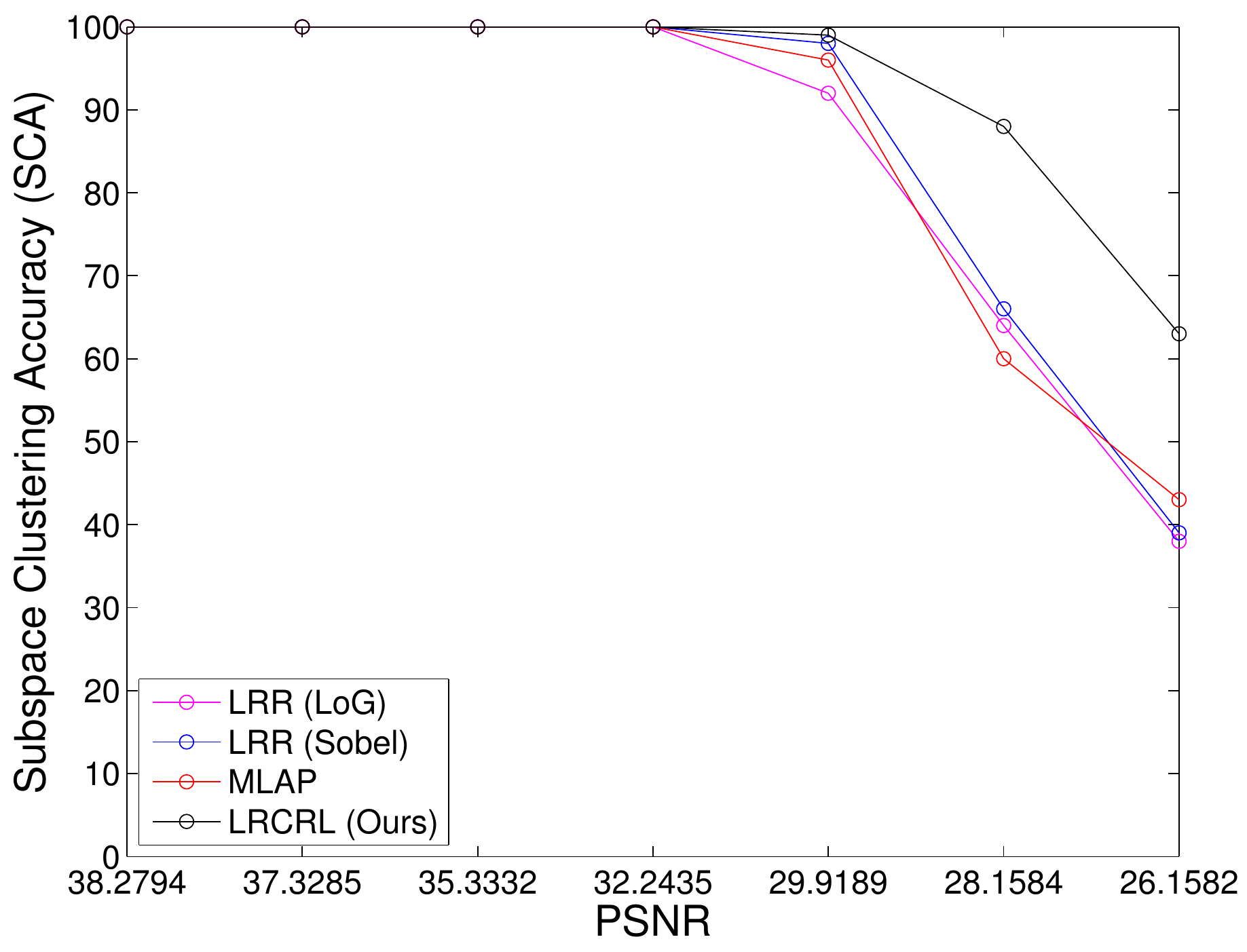}}
\caption{Results for the real data (Extended Yale Face Database B) segmentation experiment  with various magnitudes of Gaussian noise. cLRSC (ours) outperforms both MLAP and LRR at all noise levels.}
\label{Plot:FaceStats}
\end{figure*}

From this data we generated two tertiary appearances which were used for subspace clustering. The first was a Laplacian of Gaussian filtered version and the second was a Sobel edge-emphasising filtered version yielding two appearances $\{ \mathbb X_1 , \mathbb X_2 \}$. For an example of the appearance data used pleas see Figure \ref{FaceExample2}. We repeated tests 50 times for various levels of corruption with Gaussian noise. Results can be found in Figure \ref{Plot:FaceStats}. cLRSC shows an improvement in clustering accuracy over MLAP and LRR on individual appearances.

\section{Conclusion}

We presented and evaluated a novel framework for collaborative low-rank subspace clustering (cLRSC). This framework exploits the representation encoded in multiple appearances of a single phenomenon to increase discriminative power. This framework outperforms the prior state of the art in this field (LRR and MLAP). Further it is extremely robust to noise. In the future we hope to address two areas where cLRSC can be improved:
\begin{itemize}
\item Running time and memory requirements. Currently cLRSC increases the running time and memory requirements for subspace clustering when compared to LRR.
\item Strict enforcement of the rank constraint.
\end{itemize}
Recent advances in the field of Riemannian optimization \cite{Vand:2013} show promise in addressing both issues.

%

%
\subsection*{Acknowledgments}
The research project is supported by the Australian Research Council (ARC) through grant DP140102270. It is also partly supported by the Natural Science Foundation of China (NSFC) through project 41371362.

\clearpage

\bibliography{references}
\bibliographystyle{apacite}

\end{document}